\documentclass[11pt, letter]{article}
\usepackage[latin1]{inputenc}
\usepackage{amssymb,amsmath,array}
\usepackage{fullpage}
\usepackage{times}
\usepackage{helvet}
\usepackage{courier}

\usepackage{xspace}

\usepackage{algorithmic}	
\usepackage{algorithm}		
\usepackage{amsmath,multirow,amssymb,bm}

\newcommand{\bmw}{\textsf{BMW}\xspace}

\usepackage{url}

\newcommand{\OO}{\mathcal{O}}

\newcommand{\RR}{\mathbb{R}}

\newcommand{\CC}{{\bf C}}

\usepackage{graphicx}
\usepackage{calc}
\newlength{\depthofsumsign}
\newlength{\totalheightofsumsign}
\newlength{\heightanddepthofargument}

\makeatletter
\newcommand*{\DivideLengths}[2]{%
  \strip@pt\dimexpr\number\numexpr\number\dimexpr#1\relax*65536/\number\dimexpr#2\relax\relax sp\relax
}
\makeatother

\begin{document}
\title{Algebraic multigrid support vector machines}

\author{Ehsan Sadrfaridpour$^1$, Sandeep Jeereddy$^2$, Ken Kennedy$^2$, Andre Luckow$^2$\\ Talayeh Razzaghi$^1$, Ilya Safro$^1$
%
%
\vspace{.3cm}\\
%
1- Clemson University, School of Computing, Clemson SC, USA
%
\vspace{.1cm}\\
2- Innovation Lab, BMW Group IT Research Center, Information Management Americas\\ Greenville SC, USA
}

\maketitle

\begin{abstract}
The support vector machine is a flexible optimization-based technique widely used for classification problems.
In practice, its training part becomes computationally expensive on large-scale data sets because of such reasons as the complexity and number of iterations in parameter fitting methods, underlying optimization solvers, and nonlinearity of kernels. We introduce a fast multilevel framework for solving support vector machine models that is inspired by the algebraic multigrid. Significant improvement in the running has been achieved without any loss in the quality. The proposed technique is highly beneficial on imbalanced sets. We demonstrate computational results on publicly available and industrial data sets.

\end{abstract}

\section{Introduction}

Support vector machine (SVM) is one of the most well-known supervised classification methods. 
The optimal classifier is achieved through solving a convex optimization model. 
When the data is big, the training of SVM becomes highly time-consuming. One of the reasons for that is a time complexity of the underlying optimization solver required for the training. The second reason is related to finding best parameters (the model selection stage) for SVM models.
While training the classifier is a common phase in all SVMs, the model selection phase is usually applied on difficult data sets (e.g., when the data is noisy, imbalanced, and incomplete) in order to tune the parameters. On the one hand, SVM models are often much more flexible than other supervised classification methods. On the other hand, the flexibility comes with the price of finding the best model through tuning. Typically, the complexity of convex quadratic programming (QP) SVM algorithms is between $\OO(n^2)$ to $\OO(n^3)$ \cite{graf2004parallel}. For example, the solver we compare our algorithm with, namely, LibSVM \cite{chang2011libsvm}, 
which is one of the most popular QP solvers for SVM, scales between $\OO(n_{f}{n_{s}}^2)$ to $\OO(n_{f}{n_{s}}^3)$ subject to how effectively the cache is exploited in practice, where the numbers of features, and samples are denoted by $n_f$ and $n_s$ respectively. Clearly, this complexity is prohibitive for kernel based SVM models applied on practical big data without using parallelization and high-performance computing.

%

One of the major limitations of applying many standard supervised classification algorithms is the  imbalanced data, i.e., when the number of instances of one class is substantially greater than that in another class. In multi-class classification, the problem of imbalanced data is even bolder \cite{lopez2015cost}. This might dramatically deteriorate the performance of  algorithms. 
The SVM models are flexible enough to address the problem of imbalanced data. However, such models are usually computationally expensive. 
Since standard SVM algorithms often misclassify the data points of a small class, the cost-sensitive version of SVM, known as \emph{weighted support vector machine} (WSVM), has been developed. 
\emph{We are interested in developing a method that is scalable to very large data, and robust with respect to the imbalanced data}. 

In recent years, several strategies have been proposed to improve the performance of underlying QP solvers for big data. Efficient serial algorithms include decomposition techniques \cite{osuna1997improved}, shrinking and caching \cite{joachims1999making}, and fast second order working set selection \cite{fan2005working}. Another approach to accelerate the QP solvers is a chunking \cite{joachims1999making}, in which the models are solved iteratively on the subsets of training data until the global optimum is achieved. A popular LibSVM solver \cite{chang2011libsvm} implements the sequential minimal optimization (SMO) algorithm. In the cases of easier data for which kernel based SVM is not required, such approaches as LibLINEAR \cite{fan2008liblinear} exhibit good performance for linear SVMs using a coordinate descent algorithm. Another way to cope with the big data is through effective parallelization. In PSVM \cite{zhu2008parallelizing}, the algorithm reduces memory use, and parallelizes data loading and computation in interior-point solver. 
Other works utilize many-core GPUs and other architectures to accelerate SMO \cite{platt1999fast,you2015scaling}. 

In this paper, we propose a novel method for efficient and effective acceleration of (W)SVM solvers for large-scale data. In the heart of this method lies a multilevel algorithmic framework (MAF) inspired by the multiscale optimization strategies \cite{vlsicad}.
The main objective of MAF is to construct a hierarchy of problems (coarsening), each approximating the original problem but with fewer degrees of freedom. This is achieved by introducing a chain of successive restrictions of the problem domain into low-dimensional or small-size domains and solving the problem in them using local processing (uncoarsening). Typically, in computational optimization problems, the MAF combines solutions obtained by the local processing at different levels of coarseness into one global solution. Such frameworks have several key advantages that make them  attractive for applying on large-scale data: they exhibit a linear complexity,
and can be parallelized. Another advantage
of the MAF is its heterogeneity, expressed in the ability to incorporate external appropriate optimization algorithms (as a refinement) in the framework at different scales of coarseness. These frameworks are extremely successful in various practical machine learning and data mining tasks such as clustering \cite{mlmodul,KushnirGB06}, segmentation \cite{sharon06Hierarchy}, and dimensionality reduction \cite{raey}.

\noindent {\bf Our contribution}
We introduce a novel multilevel framework for fast computation of (W)SVM classifiers. The algorithm is based on the algebraic multigrid (AMG) multilevel scheme \cite{vlsicad}. We combine the AMG coarsening with the principles of: (a) coarse approximations of the support vectors, and (b) effective model selection parameter tuning through inheriting them from the coarse scales. The framework can be accelerate the performance and even improve the quality of both SVM and (W)SVM classifiers. To the best of our knowledge this is the first AMG-based algorithm for (W)SVM. The proposed method can be parallelized as any AMG algorithm, and its superiority is demonstrated on publicly available and industrial datasets of \bmw. Our work extends and generalizes previous multilevel approaches such as  \cite{razzaghi2015scalable,razzaghi2016multilevel} which results in a better running time and higher quality classifiers. 

The major difference between typical computational optimization MAF, and the (W)SVM is the output of the model. 
In (W)SVM, the main output is the set of the support vectors which is usually much smaller than the total number of data points. We use this observation in our method by redefining the training set during the uncoarsening. In particular, we inherit the support vectors from the coarse scales, add their neighborhoods, and refine the support vectors at each fine scale. 
In other words, we improve the separating hyperplane throughout the hierarchy by gradual refinement of the support vectors until a global solution at the finest level is reached.
In addition, we inherit the parameters of model selection and kernel from the coarse levels, and refine them throughout the uncoarsening.
%
%
%
%
%

\section{Support Vector Machines}
We briefly define the optimization problem underlying SVM models. 
Given $n$ data points $\{x_i\}_{i=1}^n$ in $\RR^d$, we define the corresponding labeled pairs $(x_i,y_i)$, where each $x_i$ belongs to the class determined by the given label $y_i \in \{-1, 1\}$.
 Data points with positive labels are called 
``minority'' class and are denoted by $\CC^+$, where $|\CC^+| = n^+$. The rest of the points belongs to the ``majority'' class which is 
denoted by $\CC^-$, where $|\CC^-| = n^-$.
Solving the following convex optimization problem by finding $w$, and $b$ produces the hyperplane
with maximum margin between $\CC^+$, and $\CC^-$
\begin{align}\label{eqn:SVM}
\textsf{minimize} & &  \frac{1}{2}\lVert w \rVert^2 + {C}\sum_{i=1}^n \xi_i \\
\textsf{subject to}  & & y_i (w^T \phi(x_i) + b) \geq 1 - \xi_i, &\quad i = 1, \dots,n \nonumber\\
& & \xi_i \geq 0, &\quad i=1, \dots, n \nonumber.
\end{align}
The mapping of data points to higher dimensional space is done by $\phi:\RR^d \rightarrow \RR^p ~ (d \leq p)$ to make two classes separable by a hyperplane.
The term slack variables $\{\xi_i\}_{i=1}^n$ are used to penalize the misclassified points.
The parameter $C > 0$ controls the magnitude of the penalization. 
The primal formulation is shown at (\ref{eqn:SVM}) which is known as the \textit{soft margin} SVM  \cite{wu2005svm}.
Solving the Lagrangian dual problem produces a reliable convergence which is faster than methods for primal formulation. 
The WSVM addresses imbalanced problems with assigning different weights to classes with parameters $C^+$ and $C^-$. The set of slack variables is split into two disjoint sets $\{\xi_i^+\}_{i=1}^{n^+}$, and $\{\xi_i^-\}_{i=1}^{n^-}$, respectively.
In WSVM, the objective of (\ref{eqn:SVM}) is changed into 
\begin{align}\label{eqn:WSVM}
  \textsf{minimize} ~~~~~ & \frac{1}{2}\lVert w \rVert^2 + {C^+}\sum_{i=1}^{n^+}  \xi_i^+ + {C^-}\sum_{j=1}^{n^-}  \xi_j^-. 
\end{align}
In all (W)SVM models, we use the Gaussian kernel 
$\exp (-{\gamma} \lVert x_i-x_j \rVert^2)$. Overall, in WSVM model, three parameters ($C^+$, $C^-$, and $\gamma$) require tuning which is one of the main reasons of high complexity of these solvers. 
Typically, such parameter tuning techniques (e.g., the uniform design) apply sophisticated algorithms that iteratively run the solver many times to find the optimal parameters.

\section{Algorithm}

The goal of this paper is to introduce a framework that accelerates the performance of (W)SVM solvers, while preserving or improving the quality of models. In particular, we are interested in improving the running time of nonlinear (W)SVM. However, a similar strategy is applicable with linear cases. The proposed framework is inspired by the AMG-like solvers for computational optimization problems \cite{safro2015advanced,leyffer2013fast,KushnirGB06,SafroT11}. It belongs to the family of multiscale hierarchical learning strategies with the following main phases: (a) coarsening; (b) coarsest scale learning; and (c) uncoarsening.
In the coarsening process, the original problem is gradually restricted to smaller spaces by creating aggregates of fine data points and their fractions (an important feature of AMG), and turning them  into the data points at coarse levels.  
The main mechanism underlying the coarsening phase is the AMG which successfully helps to identify the interpolation operator for obtaining a fine level solution from the coarse aggregates. When a hierarchy of coarse representations is created, and the number of coarse data points is sufficiently small, the coarsest scale learning is applied. In this stage, the (W)SVM problem is solved exactly on the coarsest aggregates.
%
In the uncoarsening phase, the solution obtained at the coarsest level (i.e., the support vectors and parameters) is gradually projected back to the finest level by interpolation and further local refinement of support vectors and parameters. A critical difference between our approach and \cite{razzaghi2015scalable} is that in our approach the coarse level support vectors are, in fact, not real data points prolongated from the finest level. Instead, they are centroids of aggregates that contain full fine-level data points and their fractions.\\
%
%
%

\noindent {\bf Framework initialization}
We initialize MAF with an undirected affinity graph $G=(V,E)$ generated from the training set of (W)SVM. Each data point $i$ is associated with node $i\in V$ (same notation is used for points and nodes), and the set $E$ is determined by the approximate $k$-nearest neighbor ($k$-NN) graph connections. We found a very little difference in the quality of the results if an exact $k$-NN graph is used while the running time for finding the approximate $k$-NN graph is significantly better. Throughout this paper, all approximate $k$-NN graphs are computed using FLANN library ~\cite{muja_flann_2009}, where $k=10$, and the distance is Euclidean. (We observed that increasing $k$ does not improve the quality of the results.) The obtained graph will serve as a structure for further coarsening.

In the multilevel graph frameworks \cite{safro:relaxml}, the edge weights represent the strength of connectivity between nodes in order to ``simulate'' the following  interpolation scheme applied at the uncoarsening phase. The stronger connection exists between two nodes, the more chances they have to interpolate a solution to each other. For the classifier learning problems, this can be expressed as a similarity measure in the spirit of \cite{KushnirGB06,fang2010multilevel}, so we define a distance function between nodes (or corresponding data points) as an inverse of the Euclidean distance in the $k$-NN graph. We omit the results of experiments with other distances which are currently being addressed in another paper as well as more advanced distance measure approaches such as \cite{brannick2006energy,ChenS11} that are often essential in multilevel methods.

In this paper, we work with binary classification problems (and one-vs-many multi-class classifiers)  but the approach is easily generalizible to multi-class classification. The coarsening is applied separately on both majority and minority classes, i.e., the $\CC^+$ points cannot be aggregated with points in $\CC^-$.\\
%
%
%

\noindent {\bf Coarsening Phase}
The main goal of the coarsening is to create a hierarchy of coarse representations of the original data manifold using the AMG coarsening applied on the approximated $k$-NN graph. We denote the sequence of $K$ next-coarser graphs by $\{G_i = (V_i, E_i)\}_{i=0}^K$, where $G_0 = G$ is the original graph that corresponds to the original training set of one of the classes, and $K$ is the number of levels in the hierarchy. For the completeness of the paper, we repeat the main steps of the AMG-based graph coarsening algorithm \cite{SafroRB06}.

We describe a two-level process of obtaining the coarse graph $G_c = (V_c, E_c)$ and the corresponding coarse training set from the current fine level $G_f = (V_f, E_f)$ and its training set (e.g., the transition from level $l$ to $l+1$). The process is started with selecting seed nodes that will serve as centers of coarse level nodes, called aggregates. Coarse nodes will correspond to the coarse data points at level $c$. Structurally, each coarse aggregate can include one full seed $f$-level point, and possibly several other $f$-level points and their fractions. Intuitively, it is equivalent to grouping nodes in $V_f$ into many small subsets allowing intersections, where each subset of nodes will correspond to a coarse point at level $c$. During the aggregation process, most coarse points will correspond to subsets of size greater than 1, so we introduce the notion of a volume $v_i\in \RR_+$ for all $i\in V$ to reflect the importance of a point or its capacity that includes finest-level aggregated points and their fractions. We also introduce the edge weighting function $w:E_l\rightarrow \RR_+$ for each graph $G_l$, $0\leq l \leq K$, to reflect the strength of connectivity and similarity between nodes.

In Algorithm \ref{alg:coarsening}, we show the details of AMG coarsening.
In the first step, we compute the future-volumes $\vartheta_i$ for all $i\in V_f$ to determine the order in which $f$-level nodes will be tested for declaring them as seeds (line 2), namely,
\begin{equation}
\label{eqn:futurevolume}
\vartheta_i = v_i + {\sum\limits_{j \in F} v_j \cdot {w_{ji} \over \sum\limits_{k \in V} w_{jk}}}.
\end{equation}
The future-volume $\vartheta_i$ is defined as a measure of how much an aggregate 
seeded by a data point $i$ (or a node in $V_f$) might potentially grow at the next level $c$. 

We assume that in the finest level, all volumes are ones. We start with selecting a dominating set of seed nodes $C\subset V_f$ to initialize future coarse aggregates. Nodes that are not selected to $C$ will belong to $F$ such that $V_f = F\cup C$. Initially, the set $F$ is set to be $V_f$, and $C=\emptyset$ since no seeds have been selected (line 1). After that, points with $\vartheta_i$ that is exceptionally larger than the  average $\overline{\vartheta}$ are transferred to $C$ as the most ``representative'' points (line 3). 
%
Then, all points in $F$ are accessed in the decreasing order of $\vartheta_i$ updating $C$ iteratively (lines 6-11), namely, if with the current $C$, and $F$, for point $i\in F$, 
\[
\sum_{j \in C} w_{ij} / \sum_{j \in V_f} w_{ij} \leq Q,
\]
where $Q$ is a threshold, i.e., the point is not strongly coupled to already selected seed points in $C$, then $i$ is moved from $F$ to $C$. 
%
%
%
Usually, the points with larger future-volumes have a better chance to be selected to $C$ to serve as centers of future coarse points.
Adding more seeds prevents too aggressive coarsening that can lead to ``overcompressed''  information at the coarse level 
and low quality classification model. However, it has been observed that in most AMG algorithms, $Q\geq 0.6$ is not required (however, this depends on the type and goals of aggregation). In our experiments $Q=0.5$, and $\eta = 2$. Other similar values do not significantly change the results.
\begin{algorithm}
  \caption{Selecting seeds for coarse nodes}\label{alg:coarsening}
  \begin{algorithmic}[1]
      \STATE $C\gets \emptyset, F\gets V_f$ \\
      \STATE {\bf Calculate} $\forall i\in F$ $\vartheta_i$, and the average $\bar{\vartheta}$\\
      \STATE $C \gets$ nodes with $\vartheta_i > \eta \cdot (\bar{\vartheta})$
      \STATE $F \gets V_f \setminus C$
      \STATE {\bf Recompute} $\vartheta_i$ $\forall i \in F$
      \STATE Sort $F$ in the descending order of $\vartheta$
      \FOR{$i \in F$}
	      \IF{$\left({\sum\limits_{j \in C} w_{ij}} / {\sum\limits_{j \in V_f} w_{ij}}\right) \leq Q$}
		      \STATE move $i$ from $F$ to $C$
	      \ENDIF
      \ENDFOR
      \RETURN $C$
  \end{algorithmic}
\end{algorithm}

When the set $C$ is selected, we compute the AMG interpolation matrix $P\in \RR^{|V_f|\times |C|}$ that is defined as
\begin{equation}\label{eq:interp}
P_{ij} =
\left \{
\begin{tabular}{cc}
$ {w_{ij}} / {\sum\limits_{k \in N_i} w_{ik}} $ & if $i \in F$, $j \in N_i$ \\
1 & if $i \in C$, $j=I(i)$\\
0 & otherwise
\end{tabular}
\right \},
\end{equation}
where $N_i = \{j \in C \mid ij\in E_f\}$ is the set of $i$th seed neighbors, and $I(i)$ denotes the index of a coarse point at level $c$ that corresponds to a fine level aggregate around seed $i\in C$. Typically, in AMG methods, the number of non-zeros in each row is limited by the parameter called the interpolation order or caliber \cite{vlsicad} (see discussion about $R$ and Table \ref{tab:ioprogr}). This parameter controls the complexity of a coarse-scale system (the number of non-zero elements in the matrix of coarse $k$-NN graph). It limits the number of fractions a fine point can be divided into (and thus attached to the coarse points). If a row in $P$
contains too many non-zero elements then it is likely to increase the number of non-zeros in the coarse graph matrix. In multigrid methods, this number is usually controlled by different approaches that measure the strength of connectivity (or importance) between fine and coarse variables (see discussion and our imlementation in \cite{SafroRB06}).


Using the matrix $P$, the aggregated data points and volumes for the coarse level are calculated.
The edge between points $p=I(i)$ and $q=I(j)$ is assigned with weight $W_{pq}^{(coarse)} = \sum\nolimits_{k \ne l} P_{ki} \cdot w_{kl} \cdot P_{lj}$.
The volume for the aggregate $I(i)$ in the coarse graph is calculated by $\sum\nolimits_{j} v_j P_{ji} $, i.e., the total volume of all points is preserved at all levels during the coarsening. The corresponding data point is defines as $\sum_j v_j P_{ji} x_j$.
%
%

The stopping criteria for the coarsening depends on the available computational resources that can be used to learn the classifier at the coarsest level. In all our experiments, the coarsening stops when the size is less than a threshold (typically, 500 points) that ensures a fast performance of the LibSVM dual solver.\\

\noindent {\bf Note:} $\blacktriangleright$ One of the major advantages of the proposed coarsening scheme is the natural ability to deal with the imbalanced data. When the coarsening is performed on both classes simultaneously, and in a small class the number of points reaches an allowed minimum, this level is simply copied throughout the rest of levels required to coarsen the big class. Since the number of points at the coarsest level is small, this does not affect the overall complexity of the framework, and the same set of points participates in the training at all next coarser levels.$\blacktriangleleft$\\

%
%
%
\noindent {\bf Coarsest Level}
When both classes are small enough, the training reinforced by the parameter tuning  is fast.
We use the uniform design (UD) as a model selection technique to tune the parameters \cite{huang2007model}. 
Another major advantage of the multilevel learning is the ability to inherit parameters $C^+$, $C^-$, and $\gamma$ during the uncoarsening. 
The tuned parameters are projected from the coarsest level back to next finer level, where they will be refined and projected up again.
The coarsest level learning is shown in Algorithm \ref{alg:coarsest}.

\begin{algorithm}
\begin{algorithmic}[1]
\caption{Coarsest level $i$ learning}\label{alg:coarsest}
  \IF {$n^+$ and $n^-$ are sufficiently small for the coarsest level}
  \STATE Calculate the best ($S_i$, $C^+_{i}$, $C^-_i$, and $\gamma_{i}$) using UD, and  (W)SVM solver on $|\CC_i^+|$ and $|\CC_i^-|$
  \STATE {\bf Return} $S_i$ (the set of coarsest support vectors), $C^+_i$, $C^-_i$, and $\gamma_i$ (learned parameters for level $i$)
  \ENDIF
\end{algorithmic}
\end{algorithm}

%
%
\noindent {\bf Uncoarsening}
When the coarsest level is solved, we start to gradually project the solution back to the finest level. In contrast to the classical multilevel methods for computational optimization problems \cite{vlsicad} in which each variable should be solved, the solution of (W)SVM consists of the set of support vectors whose size is typically much smaller than the number of data points. Thus, the main time-consuming ``operation'' of the uncoarsening is to project back and refine the set of coarse support vectors. This can be done very fast if we do not take into account all points at each level for the training. Instead, at each level, we define a new training set that includes only points from fine aggregates of the respective coarse level support vectors.

The $i+1 \rightarrow i$ uncoarsening is presented in Algorithm \ref{alg:refinement}. The set of support vectors $S_{i+1}$ and parameters $C^+_{i+1}$, $C^-_{i+1}$, and $\gamma_{i+1}$ from level $i+1$ are the inputs for level $i$. First, the new training data ($data_{\text{train}}$) is created by taking all level $i$ points from the aggregates that correspond to the support vectors in $S_{i+1}$ (lines 2-6). We denote by $I^{-1}$ the reverse index function.
%
%

The parameter tuning using UD or other similar methods is a computationally expensive part of (W)SVM training which takes most of the time for large-scale data sets. Since it can be applied at the coarse levels of small size, we verify the size of a new $data_{\text{train}}$ (parameter $Q_{dt}$), and decide whether the UD is still applicable (line 7) or not. In case it can be applied, we run it around the parameters $C^+_{i+1}$, $C^-_{i+1}$, and $\gamma_{i+1}$ inherited from the coarse level $i+1$ (lines 8-9). Otherwise, if the size of the training data is too large for the UD, we continue to inherit the parameters without adjusting them. Because in most problems, the number of support vectors is much smaller than the number of data points, even in very large data sets, we succeed to refine the parameters using UD at, approximately, 8-10 levels without any significant loss in the running time. This gives us an effective and efficient practical parameter tuning technique that has been applied for several customer satisfaction classification problems in real-world large-scale data in recommender systems of \bmw.
%

\begin{algorithm}
\begin{algorithmic}[1]
\caption{Uncoarsening from level $i+1$ to $i$}\label{alg:refinement}

  \STATE  {\bf Input:} $S_{i+1}, C^+_{i+1}, C^-_{i+1}, \gamma_{i+1}$ \\
%
\STATE $data_{\text{train}}\leftarrow \emptyset$ 
 \FORALL{$p\in S_{i+1}$} 
 \STATE $N_p \leftarrow $ all points in the aggregate $I^{-1}(p)$
 \STATE $data_{\text{train}}\leftarrow data_{\text{train}} \cup N_p$
 \ENDFOR
 
  \IF {$|data_{\text{train}}| < Q_{dt}$ }
    \STATE $C^O  \gets (C^+_{i+1},C^-_{i+1})$; $\gamma^{O} \gets \gamma_{i+1}$
    \STATE ($S_i$, $C^+_i$, $C^-_i$, and $\gamma_i$) $\leftarrow$ Run UD on (W)SVM using the initial center $(C^O, \gamma^{O})$
  \ELSE
    \STATE $C^+_i \leftarrow C^+_{i+1}$
    \STATE $C^-_i \leftarrow C^-_{i+1}$
    \STATE $\gamma_i \leftarrow \gamma_{i+1}$
    \STATE $S_i \leftarrow$ Apply (W)SVM  on $data_{\text{train}}$
  \ENDIF
  
  \STATE {\bf Return} $S_i$, $C^+_i$, $C^-_i$, and $\gamma_i$
\end{algorithmic}
\end{algorithm}
%
The framework works in a similar way for both regular SVM and WSVM. The WSVM shows better performance for classification of the small class 
when the data is imbalanced. 

\section{Computational Results}

\begin{table*}[t]
\footnotesize
\setlength{\tabcolsep}{4pt}
	\caption{Performance measures and running time (in seconds) for WSVM, and MLWSVM on publicly available data in \cite{blake1998uci}.}\label{tab:alleval}
	\vspace{-.5cm}
	\begin{center}
	\begin{tabular}{|lccccc|ccccc|ccccc|}
		\hline
		\multicolumn{6}{|c|}{Datasets} &
		\multicolumn{5}{c|}{WSVM} &
		\multicolumn{5}{c|}{MLWSVM} \\
		 Name          &  $r_{imb}$	&	 $n_f$	&  $l$	&$|\textbf{C}^+|$ & $|\textbf{C}^-|$ & ACC      & SN   & SP   & $\kappa$ & Time & ACC      & SN   & SP   & $\kappa$   & 	Time	 \\
		\hline
		Advertisement &	0.86	&	1558	&	3279	&	459	&	2820 & 0.92 &  0.99    &  0.45    & 0.67 & 231	& 0.83 & 0.92 & 0.81 & 0.86 &		213\\
		Buzz       &	0.80	&	77	&	140707	&	27775	&	112932   & 0.96  & 0.99     &  0.81    & 0.89 & 26026	& 0.88 & 0.97 & 0.86 & 0.91 &	 	233 \\
		Clean (Musk) &	0.85	&	166	&	6598	&	1017	&	5581 & 1.00 &  1.00    &  0.98    & 0.99 & 82	& 0.97 & 0.97 & 0.97 & 0.97 &		7\\
		Cod-RNA    &	0.67	&	8	&	59535	&	19845	&	39690   & 0.96 &  0.96    & 0.96   & 0.96 & 1857	& 0.94 & 0.97 & 0.92 & 0.95 &		102\\
		Forest     & 	0.98	&	54 	& 	581012	& 	9493	& 	571519   & 1.00 &  1.00	& 0.86	 & 0.92 & 353210	& 0.88 & 0.92 & 0.88 & 0.90	& 		479 \\
		Hypothyroid  &	0.94	&	21	&	3919	&	240	&	3679 & 0.99 &  1.00     &  0.75    & 0.86 & 3	& 0.98 & 0.83 & 0.99 & 0.91 &		3 \\
		Letter   &	0.96	&	16	&	20000	&	734	&	19266     & 1.00 &  1.00     &  0.97    & 0.99 & 139	&	0.98 & 1.00 & 0.97 & 0.99 &	12 \\
		Nursery   &	0.67	&	8	&	12960	&	4320	&	8640    & 1.00    & 1.00     & 1.00     & 1.00  & 192	& 1.00 & 1.00 & 1.00 & 1.00 &		2  \\
		Ringnorm   &	0.50	&	20	&	7400	&	3664	&	3736   & 0.98 & 0.99 & 0.98 & 0.98 & 26	& 0.98 & 0.98 & 0.98 & 0.98 &		2 \\
		Twonorm   &	0.50	&	20	&	7400	&	3703	&	3697    & 0.98 & 0.98     &  0.99    & 0.98 & 28	& 0.98 & 0.98 & 0.97 & 0.98 &		1	\\
		\hline
	\end{tabular}
	\end{center}
\normalsize
\end{table*}

\begin{table*}[t]
	
	\caption{Evaluation of regular and multilevel WSVM for DS1 set of \bmw benchmark.}\label{evaluation_all_bmw}
	\vspace{-.1cm}
	\begin{center}
	\begin{tabular}{|lcc|cc|cc|ccc|}
		\hline
	Class  & Size in &	Size in & \multicolumn{2}{c|}{WSVM on DS1} & \multicolumn{2}{c|}{MLWSVM on DS1} & \multicolumn{3}{c|}{MLWSVM on DS2} \\
	number & DS1  & DS2 & ACC       & $\kappa$ 	& ACC    & $\kappa$   & ACC         & $\kappa$  & Time (in sec.)\\
		\hline
		Class 1  & 6867	&	204497  & 0.87 &   0.90     &  0.79     &  0.79    & 0.80 &   0.79  &   1123  \\
		Class 2  & 373 	&       9892   & 0.99 &   0.36     &  0.90     &  0.69   & 0.63 &   0.69  &   200	   \\
		Class 3  & 5350	&       91952    & 0.96 &   0.92     &  0.91     &  0.91  & 0.83 &   0.82  &   135   \\
		Class 4  & 278	&       9339    & 0.99 &   0.42     &  0.87     &  0.57     & 0.77 &   0.71  &   52	  \\
		Class 5  & 2167	&       57478     & 0.93 &   0.62     &  0.63     &  0.69   & 0.62 &   0.66  &   53  \\
		
		\hline
	\end{tabular}
	\end{center}
\end{table*}

\begin{table*}[htb!]
\centering
\caption{Quality of classifiers on publicly available data sets for different orders of interpolation.}
\label{tab:ioprogr}
\begin{tabular}{|l|llllll|cccccc|}
\hline
Data set      &     \multicolumn{6}{c|}{$\kappa$}    &        \multicolumn{6}{c|}{Time}    \\
              & R=1   & R=2  & R=4  & R=6  & R=8  & R=10 &  R=1   & R=2  & R=4  & R=6  & R=8  & R=10\\
\hline              
Advertisement & 0.86  & 0.80 & 0.84 & 0.84 & 0.86 & 0.82 & 219  & 205 & 220 & 205 & 213 & 268 \\
Buzz          & 0.92  & 0.71 & 0.77 & 0.91 & 0.92 & 0.91 & 12   & 21  & 96  & 233 & 411 & 594 \\
Clean (Musk)  & 0.96  & 0.96 & 0.95 & 0.97 & 0.96 & 0.97 & 6    & 7   & 7   & 7   & 8   & 8 \\
Cod-RNA           & 0.94  & 0.95 & 0.95 & 0.95 & 0.95 & 0.94 & 48   & 140 & 84  & 59  & 146 & 150\\
Forest        & 0.63  & 0.51 & 0.59 & 0.90 & 0.89 & 0.85 & 84   & 68  & 168 & 479 & 1060 & 648  \\
Hypothyroid   & 0.35  & 0.58 & 0.91 & 0.76 & 0.90 & 0.77 & 1    & 1   & 2   & 3   & 4   & 4\\
Letter        & 0.97  & 0.98 & 0.99 & 0.99 & 0.99 & 0.99 & 5    & 5   & 12  & 24  & 35  & 39\\
Nursery       & 1.00  & 1.00 & 1.00 & 1.00 & 1.00 & 1.00 & 2    & 3   & 3   & 3   & 4   & 5  \\
Ringnorm      & 0.90  & 0.87 & 0.98 & 0.96 & 0.88 & 0.96 & 2    & 2   & 2   & 3   & 3   & 4  \\
Twonorm       & 0.97  & 0.97 & 0.98 & 0.98 & 0.98 & 0.98 & 2    & 1   & 1   & 1   & 1   & 2 \\
\hline
\end{tabular}
\end{table*}

 The proposed framework is implemented in C++, and PETSc library which is the collection of data structures and methods for solving scientific computing problems ~\cite{balay2014petsc}. PETSc provides a high-performance parallelization of algebraic structures that will be used in our future work that will be related to parallelization of MFA. Current implementation is not parallel. In general, based on the experience with with similar multilevel approaches \cite{vlsicad}, we anticipate the total complexity and performance of parallel version will be comparable to those of parallel AMG with small orders of interpolation. In our serial version, the linear complexity is comparable to serial AMG.
%
The data structure we use are sparse matrices and vectors in the compressed row format.
 The rest of the data structures are STL of C++ 11.
 Small-scale (W)SVM models, that appear during the refinement, are solved using LibSVM 3.20 and the approximate $k$-NN graphs are constructed using FLANN.


To evaluate our algorithms, we use sensitivity (SN), specificity (SP), G-mean ($\kappa$), and accuracy (ACC), namely,
\begin{equation}
\textrm{SN}=\frac{TP}{TP+FN}, \ \ \textrm{SP}=\frac{TN}{TN+FP}, \ \ \kappa = \sqrt{\textrm{SP}\cdot \textrm{SN}}
\end{equation}
and
\begin{equation}
 \textrm{ACC}=\frac{TP+TN}{FP+TN+TP+FN},
\end{equation}
where $TP$, $TN$, $FP$, and $FN$ are true positives, true negatives, false positives, and false negatives, respectively.

We experimented with publicly available and real-world industrial data of \bmw. The publicly available data is available at the UCI collection \cite{blake1998uci}. The industrial data of recommender system is given in two data sets, namely, DS1, and DS2. They can also be available for limited research purposes. All computational results are averages over 20 similar executions with different random seeds, and randomly reordered data. The training-test split was 80\%-20\% reinforced with $k$-fold cross validation.
%
%

In Table \ref{tab:alleval} (section ``Datasets''), we present an information about the size of the data and its split into majority and minority classes. The notation $r_{imb}$, and $n_f$ correspond to the imbalance factor, and the number of features, respectively.
%
Performance measures of regular and multilevel WSVM are presented in sections WSVM, and MLWSVM of Table \ref{tab:alleval}, respectively. Our main performance measure is $\kappa$ since we are dealing with the imbalanced classification. We observed one significant improvement in the quality of $\kappa$ in Advertisement data set. \emph{In general, on these and several other data sets, no significant difference in the quality of $\kappa$ between the proposed fast ML(W)SVM, and the full-time (W)SVM has been observed.}

The running time (in seconds) for both WSVM and MLWSVM is presented in columns ``Time'' in Table \ref{tab:alleval}. The running time includes calculation of the approximated $k$-NN graphs and UD  (model selection) for parameter tuning. \emph{We demonstrate that the proposed fast AMG inspired framework justifies the idea of multilevel algorithms for (W)SVM, and clearly exhibits superior running time.}

Not surprisingly, it is much easier to analyze benchmarks like UCI machine learning dataset than the real-life industrial data which is very noisy, and contains missing values. In the  \bmw data, there are 5 labeled classes of plain text customer satisfaction surveys. First, the plain text is converted into normalized tf-idf form using the uni-, and bi-gram information which makes the number of features approximately 200.000 because of the extensive use of the domain-specific jargon. Then, we reduce the dimensionality of the data to 100 by applying SVD projections. \emph{We note that we did not observe any change in the quality of the results for full, and reduced dimensional data except the increased running time for full dimensionality. While the multilevel (W)SVM framework running time is not fast but still realistic, the regular (W)SVM cannot be executed on such data at all without introducing significant changes such as high-performance parallelization or switching to linearized SVM version which significantly decreases the quality.}

The size of both DS1 and DS2 data sets is presented in columns 2-3, Table \ref{evaluation_all_bmw}. Different classes (1-5) correspond to different major product problems addressed in the customer satisfaction surveys.
%
For the evaluation of DS1 we focus only on the quality of the classifier because all running times are fast for this small dataset and mostly depend on the hardware, while for the DS2 set the running time is reported. While there is no loss in quality on both DS1, and DS2, the running time of MLWSVM on DS2 is substantially better than that of the regular WSVM which is measured in days, so it is comparable to the difference in running time of the Forest data set.

\noindent {\bf Does AMG help?} One of the main reasons for developing a multilevel AMG-based SVM framework was an observation that for the real data of \bmw, and experiments with complex healthcare data provided in \cite{razzaghi2016multilevel}, it is not enough to coarsen the data in the spirit of strict aggregation when data points are simply merged or eliminated based on some strong connectivity criteria such as in many clustering approaches \cite{dhillon2007weighted,fang2010multilevel}. Applying other acceleration techniques such as an ensemble SVM learning \cite{kim2003constructing,claesen2014ensemblesvm} also did not improve the quality of classifiers. We observed, that in many cases, the hyperplanes obtained at the coarse levels (i.e., without full uncoarsening) were substantially worse than the best known (but slowly computed) hyperplanes computed for the data sets that are known in the literature. Thus, we asked a question whether finding a better geometry of the data through more accurate AMG approximation of the spectral properties of the coarse approximated $k$-NN graphs can improve the quality of the classifier? We anticipated to have similar improvements to those obtained in segmentation \cite{sharon06Hierarchy}, and clustering \cite{KushnirGB06}. Unfortunately, because of several restrictions we cannot present full results of increasing interpolation order on the \bmw data but we analyze them on public data sets.

In Table \ref{tab:ioprogr}, we show the comparison of $\kappa$ for data sets from \cite{blake1998uci} for different orders of interpolation (the number of non-zeros in rows of matrix $P$, see Eq. \ref{eq:interp}). It is easy to see that for the data sets Forest, and Hypothyroid the quality of classifier is improved for increased interpolation order $R$. Improvement of the quality comes with a price of increasing running time that is demostrated in the ``Time'' section of Table \ref{tab:ioprogr}.

\noindent {\bf Omitted observations} (1) We are mostly interested in imbalanced problems, so we do not discuss the results of SVM and MLSVM, because their $\kappa$-quality is constantly worse than that of corresponding WSVM and MLWSVM. (2) We do not discuss a faster LibLINEAR solver \cite{fan2008liblinear} because of its significantly worse $\kappa$-quality. However, we note that if the data is not difficult enough, it can also be used as a part of the refinement instead of LibSVM. (3) We tested other solvers and strategies such as SVMlight \cite{joachims1999svmlight} and ensemble SVM \cite{kim2003constructing,claesen2014ensemblesvm}. While the running time of these approaches is similar, the quality of classification is worse.

\section{Conclusions}
We presented a new algorithmic framework for fast (W)SVM models. The framework belongs to the family of multiscale algorithms in which the problem is solved at multiple scales of coarseness, and gradually combined into one global solution of the original problem. We introduced the flexibility of the AMG coarsening and reinforced it with local learning of the support vectors and model selection parameters. This opens a number of interesting research directions to pursue. In particular, when the number of support vectors is indeed huge (which is not the case in many practical systems), we need to know how to combine multiple local hyperplanes into one global at the refinement stage that has to be applied locally for different clusters in the spirit of local refinement in other multiscale algorithms. Another major issue is related to effective inheritance scheme (such as bagging or ensemble SVM) of the model parameters for multiple hyperplanes. The implementation of our algorithms for ML(W)SVM is available at \cite{safro:svm:impl}.

%






\bibliographystyle{plain}
\bibliography{sigproc,paper,ilya,Talayeh}



\end{document}